\documentclass[%
 reprint,
 amsmath,amssymb,
 aps,
pr3,
]{revtex4-2}

\usepackage{multirow}
\usepackage{graphicx}
\usepackage{dcolumn}
\usepackage{bm}
\usepackage{hyperref}
\usepackage{tabularx}
\usepackage{booktabs}

\begin{document}

\title{Quantifying the Dissimilarity of Texts}

\author{Benjamin Shade} 
\author{Eduardo G. Altmann}

\affiliation{School of Mathematics and Statistics, The University of Sydney, Sydney, NSW 2006, Australia}

\begin{abstract}
  Quantifying the dissimilarity of two texts is an important aspect of a number of natural language processing tasks, including semantic information retrieval, topic classification, and document clustering. In this paper, we compared the properties and performance of different dissimilarity measures $D$ 
  using three different representations of texts---vocabularies, word frequency distributions, and vector embeddings---and three simple tasks---clustering texts by author, subject, and time period. Using the Project Gutenberg database, we found that the generalised Jensen--Shannon divergence applied to word frequencies performed strongly across all tasks, that $D$'s based on vector embedding representations led to stronger performance for smaller texts, and that the optimal choice of approach was ultimately task-dependent. We also investigated, both analytically and numerically, the behaviour of the different $D$'s when the two texts varied in length by a factor $h$. We demonstrated that the (natural) estimator of the Jaccard distance between vocabularies was inconsistent and computed explicitly the $h$-dependency of the bias of the estimator of the generalised Jensen--Shannon divergence applied to word frequencies. We also found numerically that the Jensen--Shannon divergence and embedding-based approaches were robust to changes in $h$, while the Jaccard distance was not.

  \end{abstract}


\maketitle


\section{Introduction}

Measuring the dissimilarity between texts quantitatively is a key aspect of a number of prominent natural language processing (NLP) tasks, including document matching~\cite{pham2015}, topic modelling~\cite{steyvers2007}, automatic question-answering~\cite{jiang2019}, machine translation~\cite{wang2019}, and document clustering~\cite{taghva2010}. It has also been used in a number of broader applications and empirical studies, with~examples ranging from the evaluation of the similarity and evolution of scientific disciplines~\cite{zheng2020, dias2018}, to~understanding user behaviour in social networks~\cite{tommasel2022, singh2016, tang2015}. While a number of surveys~\cite{gomaa2013, wang2020survey, vijaymeena2016, prakoso2021} highlight the plethora of available methods for capturing such dissimilarity, many studies and applications such as the ones above adopt only few dissimilarity measures (often only one), preventing them from providing useful comparisons and justification of choices. Furthermore, research involving the quantitative comparison of multiple measures will typically focus on  specific tasks~\cite{magalhaes2019, alanazi2016, boukhatem2022, upadhyay2022}, meaning that results are not generalisable to other areas and applications. In~addition, the widespread adoption of complex neural network and deep learning models for text representation has reduced interpretability, meaning that dissimilarity measures are often only evaluated based on numerical performance. These trends motivate us to explore the problem of text dissimilarity in a more holistic and task-agnostic~manner.

In order to obtain meaningful comparisons within this vast problem, we restrict our investigation to measures $D$ that are symmetric $D(\bm{p},\bm{q})=D(\bm{q},\bm{p})$, 
 positive $D(\bm{p},\bm{q})\ge 0$, with~$D(\bm{p},\bm{q})=0 \Leftrightarrow \bm{p}=\bm{q}$. This aligns with the definition of a particular class of dissimilarity functions called dissimilarity coefficients~\cite{webb2003}. We further limit our scope to measures that depend only on the two texts under consideration (i.e., $D(\bm{p},\bm{q})$ is independent of the remaining corpus). This excludes, for~instance, topic modelling approaches~\cite{blei2003}. Our focus is on a comparative study of the properties of different dissimilarity measures $D$ independent of specific tasks, since the choice of $D$ underlies many different applications, and~for each application there are multiple measures that may be appropriate. It is also worth noting that the list of possible expressions of $D$ used in our work is not intended to be exhaustive, and~further investigation should involve the use of alternative~functions.

Measuring the dissimilarity of texts depends not only on the choice of $D$ but also fundamentally on the choice of representation of the texts $\bm{p}$ and $\bm{q}$. Since there are no explicit features in text, much work has aimed at developing effective text representations. Perhaps the simplest way to represent a text is through its vocabulary, the~set of unique words present in the text. A~second approach is the bag-of-words model~\cite{salton1988}, whereby grammar and word order are disregarded, and~the text is represented as a word frequency distribution. Many empirical studies of natural language databases have found their vocabularies and word frequency distributions to exhibit certain statistical regularities. One such property is Zipf's law, which asserts that the rank-frequency distribution of words in a text can be modelled by a power law
\begin{equation}\label{eq.zipf}
f(r) \sim r^{-\gamma},    
\end{equation}
where $f(r)$ denotes the frequency of the $r$th most frequent word, and~$\gamma \geq 1$. A~second useful regularity is Heaps' law, which postulates that the number of unique words in a text, $V$, grows sublinearly as a power of the length of the text $N$ as
\begin{equation}\label{eq.heaps}
V \sim N^\beta.    
\end{equation}
It has been shown that under certain assumptions, Heaps' law can be interpreted as a direct consequence of the Zipfian rank-frequency distribution, where $\gamma = 1/\beta$ \cite{herdan1960}. 

Recent developments in the area of text representation have generally involved the use of contextual information together with simple neural network models to obtain vector space representations of words and phrases~\cite{bengio2003, mikolov2013, pennington2014}. These ideas have been extended to enable the learning of semantic vector space representations of sentences or documents. Some popular early approaches include Doc2Vec~\cite{le2014}, FastSent~\cite{hill2016}, and Word Mover's Embedding~\cite{wu2018}. Many more recent approaches are inspired by the transformer model~\cite{vaswani2017}, including Bidirectional Encoder Representations from Transformers (BERT) \cite{devlin2019} and Generative Pre-trained Transformer (GPT) \cite{radford2018}. 

In this paper, we consider three representations of texts---vocabularies, word frequency distributions, and~dense vector embeddings. For~each representation, we analyse and numerically evaluate a number of appropriate dissimilarity measures. We obtain new analytical results about estimators of $D$ and report on their dependence on both the length of documents $N$ and on the proportional difference $h$ in the length of the two texts. These results are expected to guide users on their choice of measures. While our analysis is far from exhaustive, both in terms of measures $D$ and representations, it provides a general framework and code repository that can be expanded to include new cases of~interest.

The paper is divided as follows. We start, in~Section~\ref{sec.materials}, by~introducing the dissimilarity measures $D$ and the methods used for evaluating them in specific settings. In~Section~\ref{sec.analytical}, we show our analytical calculations on the statistical properties of estimators of $D$ under a Zipfian bag-of-words model. The~numerical results obtained using the Project Gutenberg database are reported in Section~\ref{sec.numerical}. Finally, the~discussion of our main findings appears in Section~\ref{sec.discussion}. Details of our calculations and data analysis pipeline appear as Appendices, and~the code used for our analysis can be found in the repository~\cite{githubrepo}. 

\begin{table*}[!bt] 
\caption{Summary 
 of dissimilarity measures described in Section~\ref{sec.diss_measures}. $S_{\bm{p}}, S_{\bm{q}}$ denote vocabularies, $\bm{p},\bm{q}$ denote word frequency distributions, and~$u_{\bm{p}}, v_{\bm{q}}$ denote vector~embeddings.}
\label{measures_summary}
\newcolumntype{A}{>{\centering\arraybackslash}p{0.22\textwidth}}
\newcolumntype{B}{>{\centering\arraybackslash}p{0.24\textwidth}}
\newcolumntype{C}{>{\centering\arraybackslash}p{0.49\textwidth}}
\begin{tabularx}{\textwidth}{ABC}
\toprule
\textbf{Representation}	& \textbf{Name}	& \textbf{Expression}\\
\midrule
\multirow{2.5}{*}{Vocabulary}	& Jaccard distance & $D_J(S_{\bm{p}}, S_{\bm{q}}) = 1 - \frac{|S_{\bm{p}} \cap S_{\bm{q}}|}{|S_{\bm{p}} \cup S_{\bm{q}}|}$ \\
& Overlap dissimilarity & $D_O(S_{\bm{p}},S_{\bm{q}}) = 1 - \frac{|S_{\bm{p}} \cap S_{\bm{q}}|}{\min(|S_{\bm{p}}|,|S_{\bm{q}}|)}$ \\
\midrule
\multirow{2.5}{*}{Word frequency} & Jensen--Shannon divergence of order $\alpha$ & $D_{JS}(\bm{p},\bm{q}) = H\left( \frac{\bm{p} + \bm{q}}{2} \right) - \frac{1}{2}H(\bm{p}) - \frac{1}{2}H(\bm{q})$, $H_\alpha(\bm{p}) = \frac{1}{1-\alpha} \left( \sum_{i=1}^M p_i^\alpha - 1 \right)$ \\
\midrule
\multirow{3.5}{*}{Embedding} & Euclidean distance & $D_E(u_{\bm{p}}, v_{\bm{q}}) = \sqrt{\sum_{i =1}^n (u_i - v_i)^2}$ \\
& Manhattan distance & $D_M(u_{\bm{p}}, v_{\bm{q}}) =  \sum_{i=1}^n |u_i - v_i|$ \\
& Angular distance & $D_A(u_{\bm{p}}, v_{\bm{q}}) = \frac{1}{\pi}\arccos\left( \frac{u_{\bm{p}} \cdot v_{\bm{q}}}{||u_{\bm{p}}|| \: ||v_{\bm{q}}||} \right)$ \\
\bottomrule
\end{tabularx}
\end{table*}

\section{Materials and~Methods}\label{sec.materials}
\unskip

\subsection{Dissimilarity~Measures} \label{sec.diss_measures}

Suppose we have two texts, $\bm{p}$ and $\bm{q}$. Let $S_{\bm{p}}$ and $S_{\bm{q}}$ denote the vocabularies, i.e.,~the sets of unique words of~these two texts, respectively. A~common approach for quantifying the dissimilarity between such sets is the Jaccard distance:
\begin{equation}
    D_J(S_{\bm{p}}, S_{\bm{q}}) = 1 - \frac{|S_{\bm{p}} \cap S_{\bm{q}}|}{|S_{\bm{p}} \cup S_{\bm{q}}|}.
\end{equation}
A potential drawback of the Jaccard distance is that it has no sensitivity to the relative sizes of the two sets being compared. Thus, we also considered a second measure, which we referred to as overlap dissimilarity:
\begin{equation}
    D_O(S_{\bm{p}},S_{\bm{q}}) = 1 - \frac{|S_{\bm{p}} \cap S_{\bm{q}}|}{\min(|S_{\bm{p}}|,|S_{\bm{q}}|)}.
\end{equation}
Here, the~term being subtracted is often referred to as the overlap coefficient~\cite{vijaymeena2016}. 

\sloppy Next, we represented texts through their word frequency distributions. Let $\bm{p} = (p_1, p_2, ..., p_M)$ and $\bm{q} = (q_1, q_2, ..., q_M)$ denote two distributions, defined over the same set of word tokens $i = 1, 2, ..., M$. Note that this does not necessarily imply that the vocabularies of $\bm{p}$ and $\bm{q}$ are identical---there may exist words $j$ such that $p_j > 0$ but $q_j = 0$, or~vice~versa. From~an information theory perspective, a~natural measure to quantify the dissimilarity between $\bm{p}$ and $\bm{q}$ is the Jensen--Shannon divergence (JSD) \cite{lin1991},
\begin{equation}\label{jsd}
    D_{JS}(\bm{p},\bm{q}) = H\left( \frac{\bm{p} + \bm{q}}{2} \right) - \frac{1}{2}H(\bm{p}) - \frac{1}{2}H(\bm{q}),
\end{equation}
where $H$ is the Shannon entropy~\cite{cover2005}
\begin{equation} \label{entropy}
    H(\bm{p}) = -\sum_{i=1}^Mp_i \log p_i ,
\end{equation}
and $\bm{p} + \bm{q} = \sum_{i=1}^M (p_i + q_i)$. The~JSD has a number of properties that are useful for its interpretation as a distance. It is symmetric, non-negative, and~equal to 0 if and only if $\bm{p} = \bm{q}$. Furthermore, $\sqrt{D_{JS}(\bm{p}, \bm{q})}$ satisfies the triangle inequality and is thus a metric~\cite{endres2003}. Additionally, the~JSD between distributions $\bm{p}$ and $\bm{q}$ is equivalent to the mutual information of variables sampled from $\bm{p}$ and $\bm{q}$. This means that $D_{JS}(\bm{p},\bm{q})$ is equal to the average amount of information in one randomly sampled word token about which of the two distributions it was sampled from~\cite{gerlach2016}.  

In this paper, we predominately considered a generalisation of the JSD whereby $H$ in Equation~(\ref{entropy}) is replaced by the generalised entropy of order $\alpha$ \cite{havrda1967}
\begin{equation} \label{gen_entropy}
    H_\alpha(\bm{p}) = \frac{1}{1-\alpha} \left( \sum_{i=1}^M p_i^\alpha - 1 \right).
\end{equation}
This generalisation, first introduced in Ref.~\cite{burbea1982}, yields a spectrum of divergence measures $D_\alpha$ parameterised by $\alpha$. When $\alpha = 1$, we recover the usual Jensen--Shannon divergence $D_{JS}$. As~with $D_{JS}$, we have that $D_\alpha (\bm{p}, \bm{q})$ is non-negative. Furthermore, $\sqrt{D_\alpha(\bm{p}, \bm{q})}$ is a metric for any $\alpha \in (0,2]$ \cite{briet2009}. When applied to word frequency distributions, increasing (decreasing) $\alpha$ increases (decreases) the weight given to the most frequent words in the calculation of the entropy (and thus the JSD) \cite{gerlach2016}. 

Finally, we represented texts using document embeddings, which we denoted by $u_{\bm{p}}$ and $v_{\bm{q}}$, respectively. The~aim of all these approaches was to construct embeddings of texts such that semantically similar texts were close to each other in the vector space. To~do this, we used the open-access Sentence-BERT (SBERT) pretrained model \citep{reimers2019}. Specifically, we used the general-purpose all-MiniLM-L6-v2 model, a~fine-tuned version of the Microsoft MiniLM-L12-H384-uncased model \citep{wang2020minilm}. It maps to a 384-dimensional dense vector space and~was tuned on 1.17B training pairs. We used a smaller model to reduce computation time, but~we encourage the application of our techniques and code to larger and more complex embedding approaches. One alternative approach that would be of particular interest is the Longformer model~\cite{beltagy2020}.

An unfortunate shortcoming of pretrained models is that there is a limit on the size of the text that can be embedded. The~all-MiniLM-L6-v2 model has a maximum sequence length of 256 word tokens. To~create an embedding of the whole text, we divided the text into consecutive sequences each of length 256 and~computed a vector embedding of each sequence independently. These embeddings were then combined using mean pooling, whereby we took the elementwise average of all the~vectors. 

When evaluating dissimilarity using these dense embeddings, we can utilise typical methods for quantifying distance between finite-dimensional vectors. In~particular, we examine the Euclidean distance, Manhattan (taxicab) distance, and~angular distance, which is defined as the arccosine of the cosine similarity of two vectors, normalised by $\pi$ to bound values between 0 and~1.

The dissimilarity measures introduced in this section and used in this paper are summarised in Table~\ref{measures_summary}.

\begin{table*}[!tb] 
\caption{Summary of analytical results for the Jensen--Shannon divergence, where $h^*$ denotes the critical value of $h$ for which the bias of $D_\alpha$ is minimised. See Appendix~\ref{jsd_appendix} for a full~derivation. \label{jsd_tab}}
\newcolumntype{A}{>{\centering\arraybackslash}p{0.2\textwidth}}
\newcolumntype{B}{>{\centering\arraybackslash}p{0.45\textwidth}}
\newcolumntype{C}{>{\centering\arraybackslash}p{0.35\textwidth}}
\begin{tabularx}{\textwidth}{ABC}
\toprule
\textbf{Range} & $\bm{\textrm{\textbf{Bias}}[D_\alpha(\bm{\hat{p}},\bm{\hat{q}})]}$ & $\bm{h^*}$ \\
\midrule
$\alpha > 1 + 1/\gamma$ & $\frac{c\alpha}{2N}\left( \frac{1}{2h} + \frac{1}{2} - \frac{1}{h+1} \right)$ & $1 + \sqrt{2}$ \\
\midrule
$\alpha < 1 + 1/\gamma$ & $\frac{c\alpha}{2}\left( N^{-\alpha + 1/\gamma} \right) \left(\frac{1}{2} h^{-\alpha + 1/\gamma} + \frac{1}{2} - (h+1)^{-\alpha + 1/\gamma}\right)$ & $\frac{2^{\frac{\gamma}{1 - \gamma - \alpha \gamma}}}{1 - 2^{\frac{\gamma}{1 - \gamma - \alpha \gamma}}}$ \\
\bottomrule
\end{tabularx}
\end{table*}

\subsection{Data, Preprocessing, and~Analysis} \label{data_prep}

For our numerical analysis, we used the Project Gutenberg (PG) database~\cite{gutenberg}, an~online library of over 60,000 copyright-free eBooks that has been used for the statistical analysis of language for three decades. Specifically, we used the Standardised Project Gutenberg Corpus (SPGC) \cite{gerlach2020} and repository~\cite{gerlachRepo}, created by M. Gerlach and F. Font-Clos and described as ``an open science approach to a curated version of the complete PG database''. The~particular version of the PG corpus used in our research contained \mbox{55,905 books}, and~was last updated on  18 July 2018. For~reproducibility purposes, these data are available for download at \href{https://doi.org/10.5281/zenodo.2422561}{https://doi.org/10.5281/zenodo.2422560}.

A detailed description of metadata, filtering, and preprocessing can be found in \mbox{Appendix~\ref{data_description}}. An~important limitation of the metadata is that they do not include the year of the first publication of each book. As~done in Ref.~\cite{gerlach2020}, we approximated this value by assuming that all authors published their books after the age of 20 and before their death. More specifically, we said a book was published in the year $t$ if the author's year of birth satisfied $t_{birth} + 20 < t$ and the author's year of death satisfied $t < t_{death}$. 

Each text in the PG corpus has three key features that, among~many others, will affect its content and construction, namely, author, subject, and~the time period in which it was written. As~a result, we can expect that any reasonable dissimilarity measure should, in~some way, reflect these differences. Thus, we can evaluate the performance of our dissimilarity measures by determining how well they distinguish between books in the same group---i.e.,~same author, same subject, or~same time period---and books in a different~group.

Suppose we take a subcorpus of books and, using one of our dissimilarity measures, compute all dissimilarity scores between pairs of books of the same group (same author, same subject, or~same time period). These ``within-group'' values form a distribution, which we denote by the random variable $X$. In~a similar manner, let $Y$ be a random variable denoting the distribution of all the pairwise ``between-group'' dissimilarity values. If~the measure is capturing dissimilarity between texts in different groups effectively, then we would expect the within-group scores to be generally smaller than the between-group scores (this intuition is validated by the analysis in Ref.~\cite{gerlach2020}). Thus, we quantified the extent to which the dissimilarity measure depends on the grouping under consideration \mbox{by computing}
\begin{equation}\label{eq.performance}
    P(X<Y) \equiv 
    \begin{array}{c}
        \text{probability that a within-group pair has} \\
        \text{smaller $D$ than a between-group pair.}
    \end{array}
\end{equation}
Note that $0 \le P(X<Y) \le 1$, $P(X<Y) =0.5$ in the case when $D$ is independent of the grouping, and~larger values of $P(X<Y)$ correspond to better performance (stronger separation, i.e.,~less overlap, between~the distributions $X$ and $Y$). The~use of $P(X<Y)$ to evaluate the dissimilarity measures is equivalent to formulating the problem as a binary classification task ({\it Do two given texts belong to the same group?}) and using the area under a receiver operating characteristic (ROC) curve as a performance score for $D$. For~a full derivation of the relationship between the two formulations, see Appendix~\ref{roc_appendix}.

Thus, we created 30 subcorpora, 10 for each task (author, subject, and time period). Each subcorpus consisted of \mbox{1000 randomly} sampled pairs of same-group books, and \mbox{1000 pairs} of different-group books. For~each dissimilarity measure, the~quantity $P(X<Y)$ was computed on each of the 30 subcorpora. The~results of this analysis are presented and interpreted in Section~\ref{sec.numerical}, and~details about the specific texts present in each subcorpus are available in our repository~\cite{githubrepo}.

\section{Analytical~Results}\label{sec.analytical}

Suppose that texts $\bm{p}$ and $\bm{q}$ have lengths $N_{\bm{p}}$ and $N_{\bm{q}}$, respectively, where $N_{\bm{p}} \neq N_{\bm{q}}$. We are interested in examining how this difference in text length affects the measures described in Section~\ref{sec.diss_measures}. 

Now, from~an information theory perspective, we say that $\bm{p}$ and $\bm{q}$ are actually finite-size realisations of the generative processes $\bm{P}$ and $\bm{Q}$ underlying the construction of the two texts~\cite{altmann2017}. Specifically, we say that $\bm{P}$ and $\bm{Q}$ correspond to independent sampling from a Zipfian power law distribution. Thus, in~this section, we let $\bm{\hat{p}}$ and $\bm{\hat{q}}$ denote finite-size samples from this generative process. As~a result, we have that $D(\bm{\hat{p}},\bm{\hat{q}})$ is a finite-size estimator of the dissimilarity of the underlying generative processes, $D(\bm{P},\bm{Q})$. 

We investigated how the size of the samples $\bm{\hat{p}}$ and $\bm{\hat{q}}$, and~the relative difference in their sizes, affected the estimation of $D(\bm{P},\bm{Q})$. More formally, let $N_{\bm{\hat{q}}} = N$ and $N_{\bm{\hat{p}}} = hN$, where $h$ is a positive constant not equal to one. Without~loss of generality, we assumed that $h > 1$, i.e.,~that $N_{\bm{\hat{p}}} > N_{\bm{\hat{q}}}$.

\begin{table*}[!bt]
\caption{Performance of vocabulary dissimilarity measures. The~quantity $P(X<Y)$ is displayed for each measure, and~the given $p$-values are from paired $t$-tests and are~two-sided.}
\label{tab:vocab_performance}
\newcolumntype{Y}{>{\centering\arraybackslash}X}
\begin{tabularx}{\textwidth}{YYYY}
\toprule
\textbf{Task} & \textbf{Jaccard} & \textbf{Overlap} & $\bm{p}$\textbf{-Value} \\
\midrule
Author & 0.796 $\pm$ 0.013 & 0.670 $\pm$ 0.007 & $2.656 \times 10^{-9}$ \\
Subject & 0.633 $\pm$ 0.013 & 0.504 $\pm$ 0.012 & $7.153 \times 10^{-8}$ \\
Time period & 0.595 $\pm$ 0.009 & 0.478 $\pm$ 0.007 & $9.227 \times 10^{-10}$ \\
\bottomrule
\end{tabularx}
\end{table*}

\subsection{Jaccard~Distance} \label{jaccard_analytical}

By approximating the vocabulary size using Heaps' law $\left( V \sim N^\beta \right)$, we obtained the following inequality (see Appendix~\ref{jaccard_appendix} for details):
\begin{equation} \label{jaccard_bound}
    \frac{h^\beta - 1}{h^\beta + 1} \leq D_J(S_{\bm{\hat{p}}}, S_{\bm{\hat{q}}}) \leq 1.
\end{equation}
Thus, we found that the Jaccard distance $D_J(S_{\bm{\hat{p}}},S_{\bm{\hat{q}}})$ was bounded from below by an increasing function of $h$. For~simplicity, we denote this function by $g$,
\begin{equation}
    g(h) = \frac{h^\beta - 1}{h^\beta + 1} = 1 - \frac{2}{h^\beta + 1}
\end{equation}
We see that $g(1) = 0$, and~that $g(h) \to 1$ as $h \to \infty$. Interestingly, $g(h)$ is not dependent on $N$. This tells us that the lower bound of the estimator  $D_J(S_{\bm{\hat{p}}},S_{\bm{\hat{q}}})$ is not affected by the lengths of the two texts, but~only their proportional difference in~length.

In addition, we see that this lower bound makes $D_J(S_{\bm{\hat{p}}},S_{\bm{\hat{q}}})$ an inconsistent estimator of $D_J(S_{\bm{P}},S_{\bm{Q}})$. Suppose that $\bm{P} = \bm{Q}$, i.e.,~that the two texts $\bm{\hat{p}}$ and $\bm{\hat{q}}$ were sampled from the same underlying generative process. For~$D_J(S_{\bm{\hat{p}}},S_{\bm{\hat{q}}})$ to be consistent in this case, a~necessary but not sufficient condition is that $D_J(S_{\bm{\hat{p}}},S_{\bm{\hat{q}}}) \to D_J(S_{\bm{P}},S_{\bm{Q}}) = 0$, since $D_J(S_{\bm{P}},S_{\bm{Q}}) = D_J(S_{\bm{P}},S_{\bm{P}}) = 0$. However, $g(h) > 0$ if $h >1$, implying that $D_J(S_{\bm{\hat{p}}},S_{\bm{\hat{q}}}) > 0$. Hence, if~the two texts are of unequal lengths, the~estimator $D_J(S_{\bm{\hat{p}}},S_{\bm{\hat{q}}})$ is strictly larger than zero, even if the underlying generative processes are identical and $N \to \infty$. Thus, $D_J(S_{\bm{\hat{p}}},S_{\bm{\hat{q}}})$ is not a consistent estimator of the underlying $D = 0$ dissimilarity of the~vocabularies.

The overlap dissimilarity is not restricted by such a bound. It is straightforward to show that $D_O(S_{\bm{\hat{p}}},S_{\bm{\hat{q}}}) = 0$ if either $S_{\bm{\hat{p}}} \subseteq S_{\bm{\hat{q}}}$ or $S_{\bm{\hat{q}}} \subseteq S_{\bm{\hat{p}}}$, and~that $D_O(S_{\bm{\hat{p}}},S_{\bm{\hat{q}}}) = 1$ if $S_{\bm{\hat{p}}}$ and $S_{\bm{\hat{q}}}$ are~disjoint.

\subsection{Jensen--Shannon~Divergence} \label{jsd_analytical}

In Ref.~\cite{martinphd}, the~bias of the estimator $D(\bm{\hat{p}},\bm{\hat{q}})$ is computed for texts of identical length. We extended this analysis by generalising it to allow for unequal sample sizes (specifically, when $N_{\bm{\hat{q}}} = N$ and $N_{\bm{\hat{p}}} = hN$, as~stated previously).

As in Ref.~\cite{martinphd}, we see that the expression of the bias varies depending on whether $\alpha$ is larger or smaller than $1+1/\gamma$ (recall that $\gamma$ is the exponent of the Zipfian word-frequency distribution). In~both cases, we found that the bias was a decreasing function of $h$ when $h \in (1, h^*)$, and~increasing when $h > h^*$. This means that for a fixed $N$, there exists an optimal relation between text lengths, $h^*$, for~which the bias is minimised. The~bias formulations and corresponding $h^*$ values are displayed in Table~\ref{jsd_tab}, with~a full derivation presented in Appendix~\ref{jsd_appendix}.

From the above formulations, we can also observe how the bias decays as the text length $N$ grows. When $\alpha > 1 + 1/\gamma$, the~decay of the bias is $1/N$, and~$\textrm{Bias}[D_\alpha(\bm{\hat{p}},\bm{\hat{q}})] \to 0$ as $N \to \infty$. When $1/\gamma < \alpha < 1 + 1/\gamma$, the~bias again decays to zero, but~does so sublinearly. Finally, when $\alpha < 1/\gamma$, the~bias diverges as $N \to \infty$, and~thus, the~estimator $D_\alpha(\bm{\hat{p}},\bm{\hat{q}})$ also~diverges.

\section{Numerical~Results}\label{sec.numerical}

\begin{figure*}[!bt] 
    \includegraphics[width=0.8\textwidth]{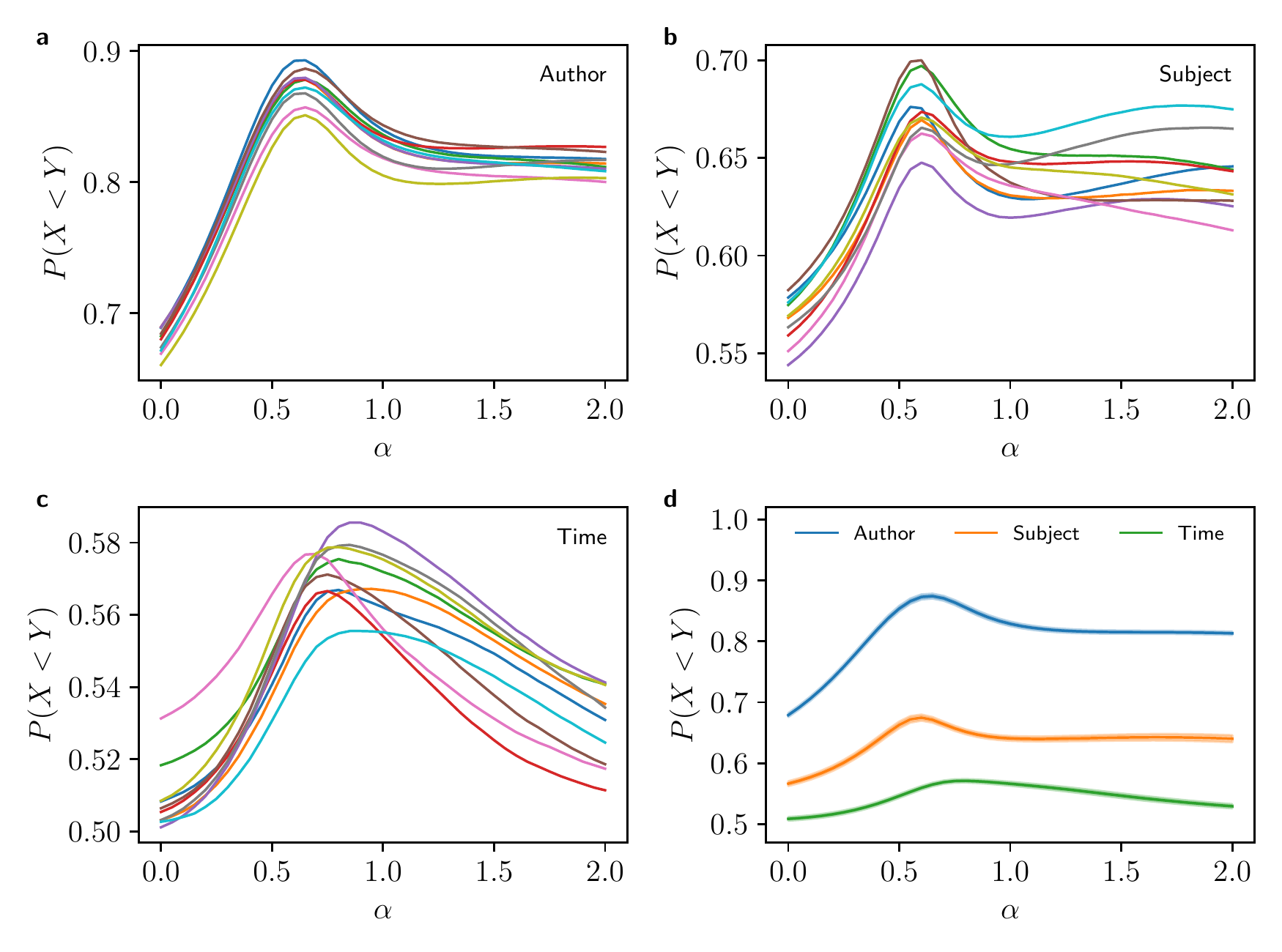}
    \caption{Performance $P(X<Y)$ (Equation~(\ref{eq.performance})) of the Jensen--Shannon divergence for different values of the parameter $\alpha$ (as defined in Equation~(\ref{gen_entropy})). (\textbf{a}--\textbf{c}) Results for author, subject, and~time period tasks, respectively. Each line shows the performance obtained on a unique subcorpus, see Section~\ref{data_prep} for details. (\textbf{d}) Performance comparison between tasks, with~the lines indicating average performance across the 10 subcorpora and shaded regions indicating standard~error.}
    \label{fig:opt_alph_uncontrolled}
\end{figure*}

\begin{table*}[!bt]
\caption{Performance comparison of the generalised JSD for $\alpha = 1$ and $\alpha = \alpha^*$. The~quantity $P(X<Y)$ is displayed for each measure, and~the given $p$-values are from paired $t$-tests and are~two-sided.}
\label{tab:alpha1_alphastar_comp}
\newcolumntype{Y}{>{\centering\arraybackslash}X}
\begin{tabularx}{\textwidth}{YYYY}
\toprule
\textbf{Task} & \textbf{JSD,} $\bm{\alpha = 1}$ &\textbf{ JSD,} $\bm{\alpha = \alpha^*}$ & $\bm{p}$\textbf{-Value} \\
\midrule
Author & 0.8288 $\pm$ 0.0035 & 0.8743 $\pm$ 0.0038 & $1.901 \times 10^{-10}$ \\
Subject & 0.6408 $\pm$ 0.0038 & 0.6749 $\pm$ 0.0048 & $1.862 \times 10^{-5}$ \\
Time period & 0.5664 $\pm$ 0.0030 & 0.5712 $\pm$ 0.0026 & $1.630 \times 10^{-2}$ \\
\bottomrule
\end{tabularx}
\end{table*}

We begin with a numerical performance comparison of our dissimilarity measures, before~conducting an investigation into the effect of varying text~lengths.

\subsection{Numerical Performance~Comparison} \label{numeric_perf_comparison}

\subsubsection{Vocabularies}

We compared the Jaccard distance and overlap dissimilarity, and~found that the Jaccard distance led to significantly higher $P(X<Y)$ values in all three tasks ($p$-value $<10^{-7}$, using two-sided paired $t$-tests). Based on these results, we chose to focus on the Jaccard distance in the subsequent analysis. These results are displayed in Table~\ref{tab:vocab_performance}.

\subsubsection{Word~Frequencies}

Since the choice of parameter $\alpha$ in the generalised Jensen--Shannon divergence affects how different frequency ranges are weighted, we sought to identify which value optimised performance across our three tasks. Figure~\ref{fig:opt_alph_uncontrolled}a--c show the performance of the generalised JSD across the parameter space $\alpha \in [0, 2]$ for each of our three tasks. All three plots indicate a clear global maximum, at $\alpha = \alpha^* = 0.65$ for the author task, $\alpha^* = 0.6$ for subjects, and~$\alpha^* = 0.8$ for time periods. The~curves also exhibit largely similar structure, with~the main notable exception being the emergence of a second, local maximum in the region $\alpha \in [1.4, 2]$ for some iterations of the subject task evaluation. Figure~\ref{fig:opt_alph_uncontrolled}d compares the performance between~tasks.

It is worth noting the difference in performance between the optimal $\alpha$ values, denoted by $\alpha^*$, and~the value $\alpha = 1$, which recovers the standard JSD in Equation~(\ref{jsd}). Table~\ref{tab:alpha1_alphastar_comp} provides the $p$-values for the paired $t$-tests between the $P(X<Y)$ calculations for the two $\alpha$ values. When using the Bonferroni adjusted threshold $0.05/3 = 1.6 \times 10^{-2}$, we see that all three results are significant, acknowledging that the significance for the time period task is marginal. These results indicate the value of considering entropies other than the popular and widely used Shannon entropy when using the Jensen--Shannon divergence, as~it may potentially lead to improved performance depending on the~application.

A surprising result in the above analysis is that the optimal performance occurred at very similar values of the parameter $\alpha$. We might have expected that different ranges on the frequency spectrum would have been emphasised for the different tasks. For~example, the~distribution of high-frequency words may have indicated stylometry and thus authorship variations, while low-frequency keywords may have been more useful for distinguishing between subjects. Our results, however, suggest that in general, penalising common words and emphasising low-frequency words improves our ability to distinguish between documents. This aligns with Ref.~\cite{mikolov2013}, where the authors found that subsampling common words led to practical benefits, including accelerated model learning and an improved accuracy of the learned vectors of rare words. Their heuristic subsampling strategy---Equation~(5) in Ref.~\cite{mikolov2013}---resulted in an effective reduction of the difference between the frequency of words, with~words with frequency $f$ appearing with frequency proportional to $\sqrt{f}$ after subsampling. This is effectively what is achieved using JSD with $\alpha<1$ (with $\alpha=0.5$ reproducing their square-root heuristic) and our finding of $\alpha^* <1$ can be seen as an information-theoretic justification for this proposed~heuristic. 

Another possible explanation for the position of $\alpha^*$ is the influence of the critical value of $\alpha$ discussed in Ref.~\cite{altmann2017}. There, it is shown that if word frequencies follow Zipf's law, then for all $\alpha < \alpha_c = 1/\gamma$, $H_\alpha$ and $D_\alpha$ diverge as the vocabulary size increases. When using the parameter estimates found in Ref.~\cite{gerlach2013}, we obtained $\alpha_c \approx 1/1.77 \approx 0.56$, which was close to our optimal $\alpha^*$ values. Due to the finite number of words in our database, we could not increase the vocabulary size without bound, and~thus we did not empirically observe $H_\alpha, D_\alpha \to \infty$. However, there was still a finite-size effect that depended on the lengths of the texts being used, and~this may have been influencing our~results. 

A third factor potentially causing the surprising similarity between our optimal $\alpha$ values is a confounding between the three different features---author, subject, and~time period. To~investigate this, we repeated our analysis on controlled subcorpora that sought to mitigate confounding. For~the author task, each corpus was limited to one subject within a 50-year period. For~distinguishing between subjects, we again limited to a 50-year period, and~for the time period tasks, we restricted the corpora to one subject (see our repository~\cite{githubrepo} for further details about these filtered subcorpora). We found that controlling the subcorpora led to a much greater variability in the curves of $\alpha$ against $P(X<Y)$, with~local maximums occurring at different points on the $\alpha$ spectrum. Importantly, however, we found that when we averaged over our 10 subcorpora, the~peaks were very close to those we computed on the uncontrolled corpora. We found that for the author task, the~maximum occurred at $\alpha^* = 0.7$, while the strongest performance for both the subject and time period tasks was at $\alpha^*=0.6$.

\begin{table*}[!bt]
\caption{Performance comparison of vector embedding approaches. The~quantity $P(X<Y)$ is displayed for each measure, and~the given $p$-values are from paired $t$-tests and are~two-sided.}
\label{tab:embedding_comp}
\newcolumntype{Y}{>{\centering\arraybackslash}X}
\begin{tabularx}{\textwidth}{YYYY}
\toprule
\textbf{Task} & \textbf{Euclidean} & \textbf{Manhattan} & \textbf{Angular} \\
\midrule
Author & 0.8448 $\pm$ 0.0026 & 0.8447 $\pm$ 0.0026 & 0.8599 $\pm$ 0.0025 \\
Subject & 0.6847 $\pm$ 0.0042 & 0.6846 $\pm$ 0.0043 & 0.6966 $\pm$ 0.004 \\
Time period & 0.5354 $\pm$ 0.0034 & 0.5357 $\pm$ 0.0034 & 0.5293 $\pm$ 0.0035 \\
\bottomrule
\end{tabularx}
\end{table*}

\begin{table*}[!bt]
\caption{Two-sided, paired $t$-tests comparing the performance of embedding distance~measures.}
\label{tab:embedding_ttests}
\newcolumntype{Y}{>{\centering\arraybackslash}X}
\centering
\begin{tabularx}{\textwidth}{YYYY}
\toprule
\textbf{Task} & \textbf{Metric Pair} & $\bm{p}$\textbf{-Value} \\
\midrule
 & Euclidean--Manhattan & $0.3185$ \\
Author & Angular--Manhattan & $1.316 \times 10^{-8}$ \\
& Angular--Euclidean & $2.002 \times 10^{-8}$ \\
\midrule
& Euclidean--Manhattan & $0.4101$ \\
Subject & Angular--Manhattan & $4.065 \times 10^{-6}$ \\
& Angular--Euclidean & $3.110 \times 10^{-6}$ \\
\midrule
& Euclidean--Manhattan & $0.3221$ \\
Time period & Angular--Manhattan & $2.202 \times 10^{-4}$ \\
& Angular--Euclidean & $3.972 \times 10^{-4}$ \\
\bottomrule
\end{tabularx}
\end{table*}

\subsubsection{Embeddings}

Across all three tasks, the~difference in performance between the Euclidean and Manhattan distances was not statistically significant, with~$p$-value $> 0.3$ for all three paired $t$-tests. The angular distance led to a stronger performance in the author and subject tasks, and~this improvement was statistically significant ($p$-value $< 10^{-5}$ for comparisons with both Manhattan and Euclidean distances). Conversely, the angular distance resulted in a significantly reduced performance in the time period task ($p$-value $< 10^{-3}$ when comparing with either Manhattan or Euclidean distance). Thus, we conclude that the angular distance is the optimal distance measure for distinguishing between both authors and subjects. The Manhattan distance was optimal for the time period task, but~its average performance increase over the Euclidean distance was marginal ($0.5357 \pm 0.0034$ compared to $0.5354 \pm 0.0034$). The~full results can be found in Table~\ref{tab:embedding_comp}, and~the $p$-values described above are available in Table~\ref{tab:embedding_ttests}.

\subsubsection{Overall~Comparison} \label{overall_comp}

\begin{figure*}[!bt] 
    \includegraphics[width=0.7\textwidth]{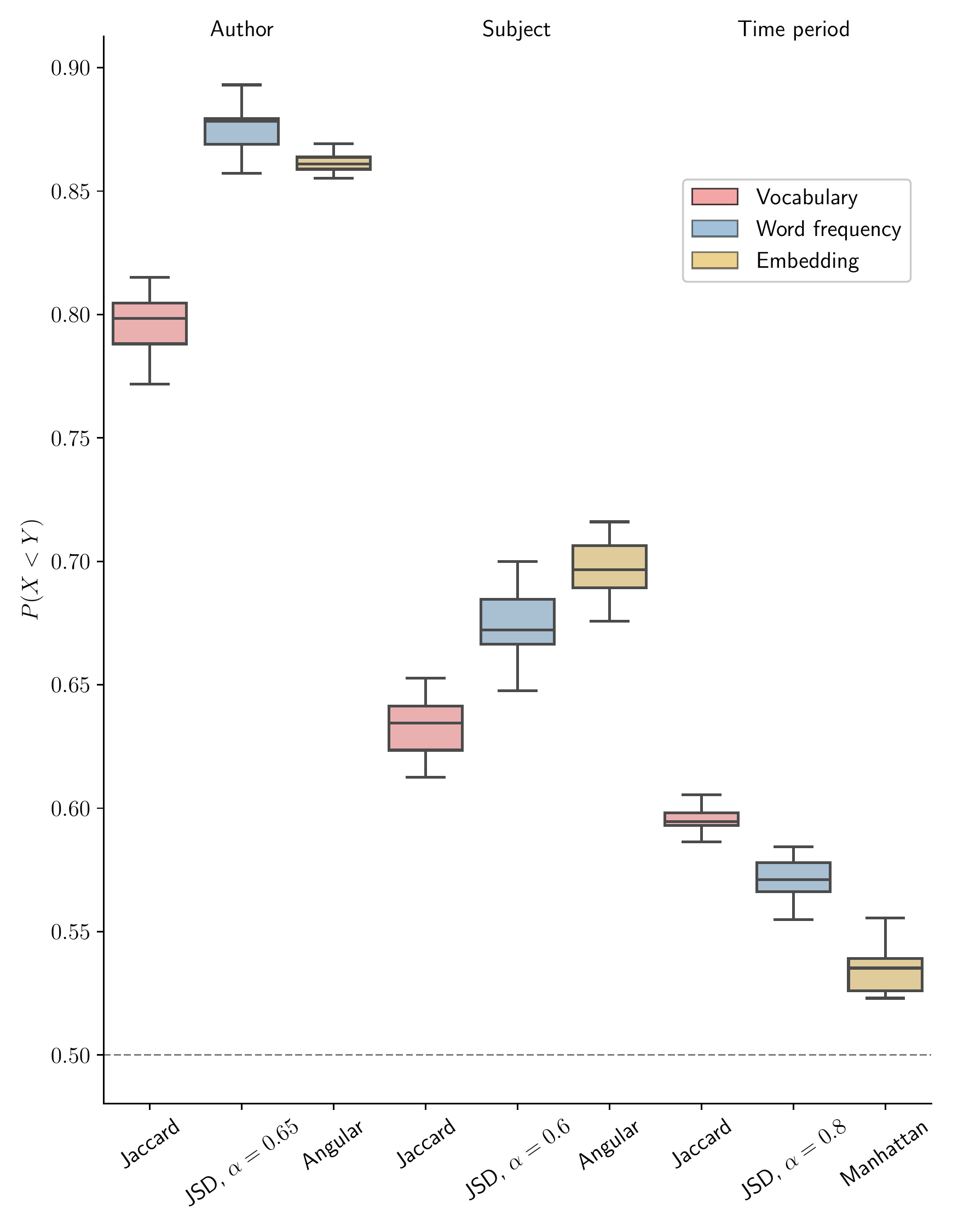}
    \caption{Performance $P(X < Y)$ (Equation~(\ref{eq.performance})) of strongest measures for each text representation across the three tasks. The~optimal measures for each text representation (vocabulary, word frequency, and~embedding) and task (author, subject, time) are identified in Section~\ref{overall_comp}. The~boxplots show the distribution of $P(X<Y)$ values for the 10~subcorpora.}
    \label{fig:final_comparison}
\end{figure*}

Figure~\ref{fig:final_comparison} displays the performance of the optimal measures for each representation across our three tasks. Here, we see that the generalised Jensen--Shannon divergence with optimal parameter $\alpha = 0.65$ had the strongest performance when distinguishing between texts written by different authors. Furthermore, the~difference in performance between the optimal JSD and the second-strongest dissimilarity measure, the angular distance, was statistically significant ($p = 2.951 \times 10^{-4}$). The Jaccard distance, which compares vocabularies, showed the weakest performance. It is perhaps surprising that a measure based on the simple bag-of-words model outperformed those based on vector embeddings. This suggests that the choices of words and their usage throughout a text are particularly useful for identifying stylistic differences between authors. It may also suggest that our vector embeddings were not fully capturing certain structural aspects of the texts that may be helpful for distinguishing between authors. Furthermore, we see that just comparing vocabularies is not sufficient; we also need to understand the distribution of those words throughout the~text.

\begin{figure*}[!bt]
    \includegraphics[width=0.7\textwidth]{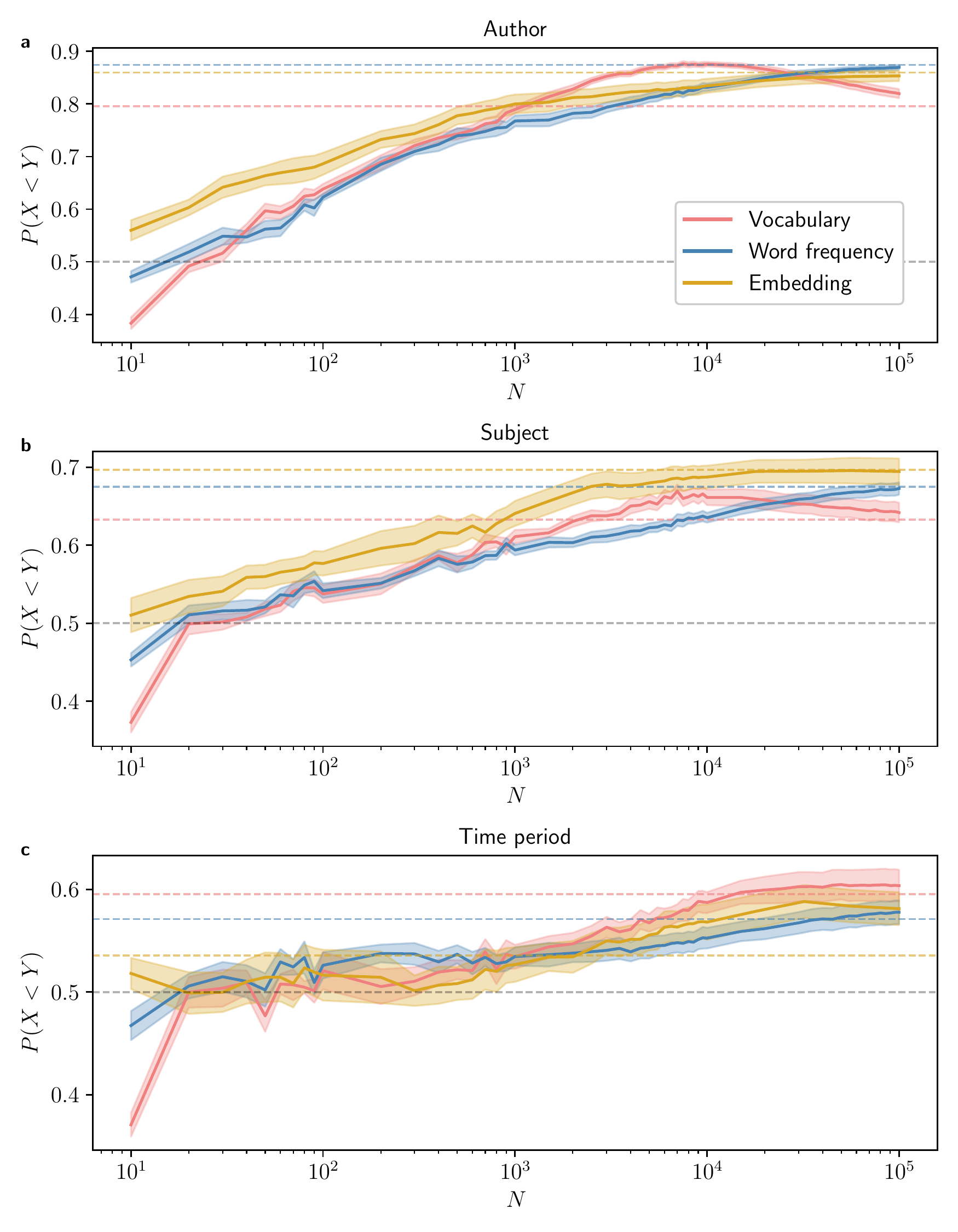}
    \caption{Performance $P(X < Y)$ (Equation~(\ref{eq.performance})) of our optimal measures as a function of text length $N$. (\textbf{a}) Author task;  (\textbf{b}) subject task; (\textbf{c}) time period task. The~optimal measures for each representation and task are identified in Section~\ref{overall_comp} and displayed in Figure~\ref{fig:final_comparison}. The~solid lines indicate average performance across the 10 subcorpora, and~the shaded regions indicate standard error. The~dashed lines indicate the average performance of these measures computed using the original texts (see Section~\ref{numeric_perf_comparison}), and~the grey line indicates the performance baseline of 0.5. In~each iteration, texts were resampled to a fixed length $N$ before computing $P(X<Y)$.}
    \label{fig:resampling_final}
\end{figure*}

When evaluating the ability of our measures to distinguish between subjects, we see that the approaches based on vector embeddings were the strongest performers. This advantage over the second-strongest performer, the JSD with optimal parameter $\alpha = 0.6$, was again statistically significant ($p = 6.665 \times 10^{-4}$). As~in the author task, we see that the measures based on vocabularies had the weakest performance. It is worth noting, however, that the average performance of the Jaccard distance ($0.6326 \pm 0.0041$) was not far below that of the standard JSD ($0.6408 \pm 0.0038$). The~difference between them was still statistically significant ($p = 1.306 \times 10^{-3}$), but~this significance was marginal if we adjusted for multiple testing. Thus, we see that the embedding approach was the best approach for capturing textual differences between subjects. Clearly, a better representation of the semantic meaning of a text, rather than just the chosen words, leads to an improved ability to identify topical differences. Furthermore, vector embeddings are able to identify words and phrases that are synonymous, while the bag-of-words model~cannot.

Rather surprisingly, the~Jaccard distance had the strongest performance when distinguishing between time periods, despite its relative simplicity. Furthermore, the~difference in performance between the Jaccard distance and the second-strongest measure, the JSD with $\alpha = 0.8$, was highly significant, with~a $p$-value of $5.374 \times 10^{-8}$. The~measures based on vector embeddings performed relatively poorly and~were only just above the baseline of 0.5. Languages change, develop, and grow over time, so it is reasonable to expect that the vocabulary of a text written today will differ from that of one written 200 years ago. What is surprising in these results, however, is that the JSD, which captures both vocabularies and word frequencies, underperformed the Jaccard distance. The~poor performance of the embedding approaches suggests that while language may change over time, similar ideas and topics are being conveyed, and~thus semantic meaning is not as useful for identifying temporal~differences. 

Additionally, Figure~\ref{fig:final_comparison} visualises the inherent difficulty of each of our three tasks. Distinguishing between authors was the easiest of the three, with~performance values typically ranging between 0.75 and 0.9. By~comparison, it was very challenging for any of our dissimilarity measures to distinguish effectively between books written in different time periods. For~this task, $P(X<Y)$ rarely exceeded 0.6, and~was typically close to the baseline of 0.5. This underperformance is likely due to the overlap between the time periods caused by our estimation of the publication date. For~the task of distinguishing between different subjects, we see that $P(X<Y)$ typically lay between 0.6 and~0.75. 

\subsection{Dependence on Text~Length}

We first observed the impact of text length on performance, with~the aim of determining the optimal measure for particular ranges of $N$. For~each task, we selected the optimal dissimilarity measure for each of the three text representations---vocabularies, word frequencies and embeddings. For~details of which measures were selected, refer to Section~\ref{numeric_perf_comparison}. 

In Figure~\ref{fig:resampling_final}, we see that in both the author and subject tasks, the~embedding approach led to a stronger performance for small text lengths, i.e.,~when $N \in [10^1,10^3]$. Thus, while our vocabulary or frequency measures may be appropriate for quantifying dissimilarity between large texts such as books, embedding approaches are more suitable for comparing short texts such as tweets or~articles.

Interestingly, in~both the author and subject tasks, the~performance of the Jaccard distance appeared to peak at around $N = 10^4$. When generating a sample of size $N$, we generally expect to sample words that have frequency $1/N$ or greater. Thus, when resampling vocabularies, $N$ is effectively a frequency cutoff---words with frequency greater than $1/N$ are likely to enter the vocabulary, while those with frequency less than $1/N$ likely will not. The~peaks in our results seem to suggest that there exists an optimal $N^*$ such that if we only include words in our vocabulary with a frequency greater than $1/N^*$, our performance will be~maximised.

The results on the time period task were less conclusive. We see that there was no clear optimal approach when comparing small texts. Importantly, the~performance of all three approaches was very close to the baseline of 0.5. While this may indicate the innate difficultly of distinguishing between time periods when comparing small texts, it is likely also resulting from our approximation of publication~date. 

\subsection{Impact of Unequal Text~Length}

\begin{figure*}[!bt] 
    \includegraphics[width=0.8\textwidth]{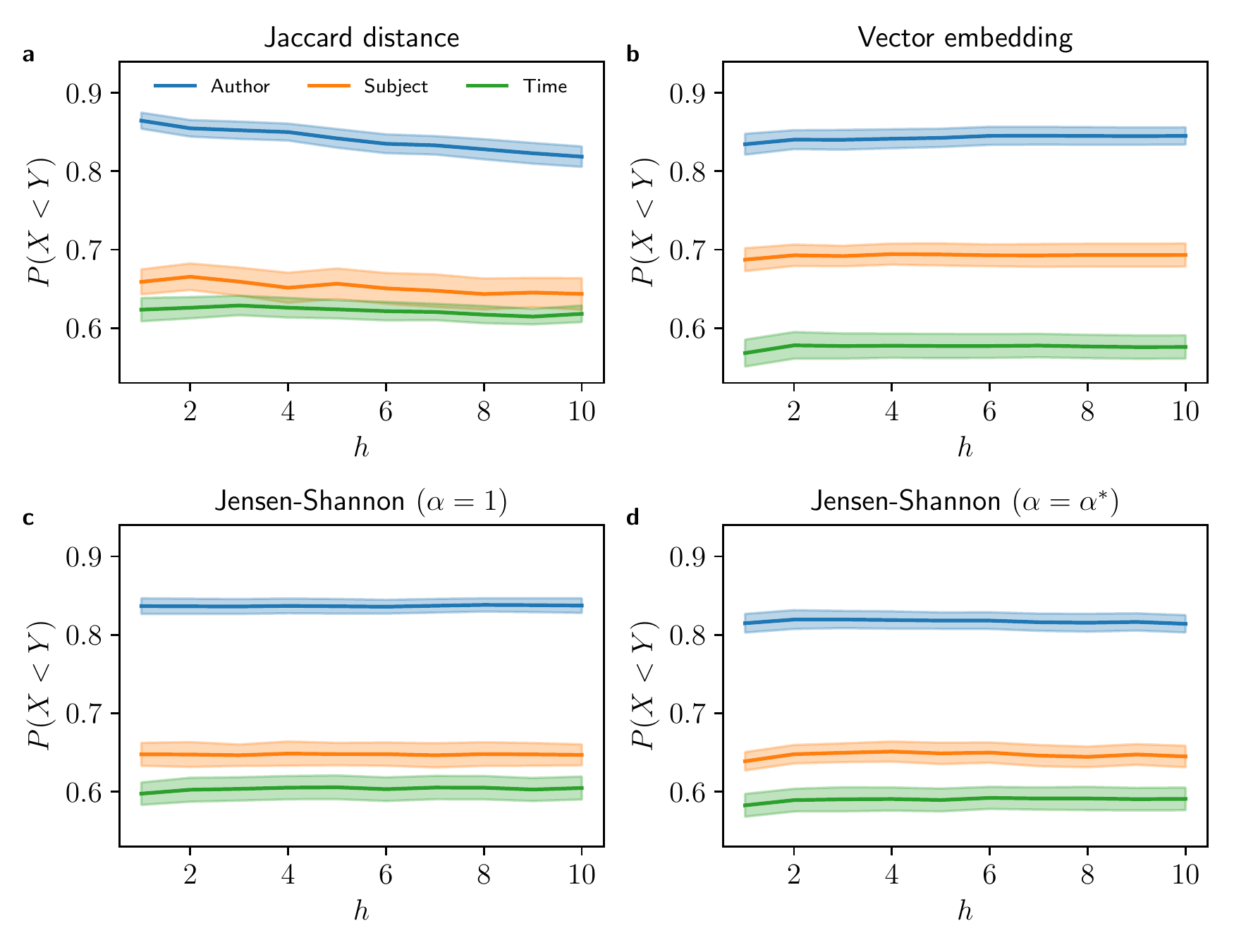}
    \caption{Impact of $h$ parameter (proportional difference in text length) on performance $P(X<Y)$ (Equation~\ref{eq.performance})) of dissimilarity measures. (\textbf{a}) Jaccard distance; (\textbf{b}) optimal embedding measure for each representation; (\textbf{c}) standard JSD based on Shannon entropy; (\textbf{d}) generalised JSD with $\alpha = \alpha^*$. For~each dissimilarity calculation, one text is resampled to length $N$, and~the other to length $hN$, where $N =$ 10,000 is fixed. The~solid lines indicate average performance across the 10 subcorpora, and~the shaded regions indicate standard~error.}
    \label{fig:h_test}
\end{figure*}

From Figure~\ref{fig:h_test}, we see that the JSD and embedding-based approaches were robust against unbalanced text lengths, with~performance remaining almost unchanged across all chosen $h$ values (recall from Section~\ref{sec.analytical} that $N_{\bm{q}} = N$ and $N_{\bm{p}} = hN$). The~finding regarding the JSD may be connected to the bias computed analytically in Section~\ref{jsd_analytical}---while the bias is dependent on $h$, it is negligible for large values of $N$, and~thus the impact of unbalanced text length should be minimal. In~contrast, the Jaccard distance was not robust, with~all three representations showing a decreasing performance with increasing values of $h$. Again, this may be related to our analytical result in Section~\ref{jaccard_analytical} regarding the lower bound of the Jaccard distance. Due to this bound, books that vary significantly in text length will receive high dissimilarity scores regardless of how fundamentally different they are, which in turn will influence the calculation of $P(X<Y)$. However, in~both cases, further investigation is required to fully establish~causality.

Now, a~possible way to extend the Jensen--Shannon divergence to account for unequal text lengths is by giving different weightings to the two distributions being \mbox{compared~\cite{lin1991, martinphd}}. Specifically, we say
\begin{equation}
    D_\alpha(\bm{p}, \bm{q}) = H_\alpha(\pi_{(\bm{p})} \bm{p} + \pi_{(\bm{q})} \bm{q}) - \pi_{(\bm{p})}H_\alpha(\bm{p}) - \pi_{(\bm{q})}H_\alpha(\bm{q}),
\end{equation}
where $\pi_{(\bm{p})} = N_{\bm{p}}/(N_{\bm{p}} + N_{\bm{q}})$ and $\pi_{(\bm{q})} = N_{\bm{q}}/(N_{\bm{p}} + N_{\bm{q}})$ (recall that $N_{\bm{p}}$ and $N_{\bm{q}}$ denote the lengths of texts $\bm{p}$ and $\bm{q}$, respectively). We found, however, that the use of these weightings led to a decreased performance across all $\alpha \in [0,2]$ for all three tasks. This result is likely explained by the variability in text length in our PG data; if $N_{\bm{p}} \gg N_{\bm{q}}$, then $\pi_{(\bm{p})} \approx 1$ and $\pi_{(\bm{q})} \approx 0$, meaning that $D_\alpha(\bm{p}, \bm{q}) \approx 0$. Thus, when comparing a large text with a comparatively small text, the~weighted JSD will approach zero regardless of how different the two word frequencies~are.

\section{Discussion}\label{sec.discussion}

A number of key results emerged from our numerical performance comparison of the chosen dissimilarity measures. One such observation was the consistently strong performance of the Jensen--Shannon divergence (JSD) across the three tasks. Importantly, we saw an improvement in performance over the standard JSD  when using the generalised divergence $D_\alpha$ based on the $\alpha$-entropy (see Equation~(\ref{gen_entropy})), with~the best results in all three tasks obtained when $\alpha \in [0.6, 0.8]$. As~with the standard JSD (recovered when $\alpha=1$), $D_\alpha$ can be written as a sum over word types, thus providing a clear interpretation of the dissimilarity between texts. Other potentially important aspects of this family of dissimilarity measures are that they satisfy the triangular inequality for $\alpha \in [0,2]$~\cite{briet2009}, and~that the properties of their estimators can be computed explicitly for the simple bag-of-words process, as~done previously in Ref.~\cite{martinphd} and extended here for the case of texts with different lengths. Further studies, particularly into how the parameter $\alpha$ weights particular groups of words on the word frequency spectrum, are needed to fully understand the appropriateness of different parameter choices for particular tasks and~to interpret our finding that $\alpha<1$ is typically preferred (higher weight to low-frequency words~\cite{gerlach2016}).

Ultimately, the~choice of dissimilarity measure depends on the task, the~texts being compared, the~statistical and mathematical properties of the measures and their estimators, and~the extent to which it is important to have interpretable dissimilarity measures. For~instance, while the Jaccard distance between vocabularies is arguably the measure with the simplest interpretation, we found analytically and empirically that the effectiveness of the Jaccard distance was hindered when the two texts varied significantly in length (while, in~contrast, the~generalised JSD and embedding-based distances were all robust). Focusing only on performance, the~results we obtained indicate that the optimal dissimilarity measure is task-dependent, with~all three approaches---vocabularies, word frequency distributions, and~vector embeddings---exhibiting optimal performance on a particular task. In~addition, the~length of texts in the corpus must be considered; measures based on vector embeddings consistently outperformed other approaches on small texts, with~a likely explanation for this being that small samples were not sufficiently representative of the underlying vocabularies or word frequency distributions. When deciding on the appropriateness of a particular measure for research or application, one should weigh up the performance increase of embedding approaches with the more limited interpretability and control of statistical properties when compared to vocabulary and word frequency~techniques.

Further research in this area would involve extending our analysis to other dissimilarity measures, and~exploring additional properties, both analytically and numerically, of~such measures. Through this more holistic and standardised approach, researchers may gain a better understanding of why particular measures are suitable for certain tasks, as~well as which measures are most applicable for new and emerging NLP areas. Our work may also help to inform decisions made by industry or other users who may not have the time or resources to conduct extensive~comparisons.

\section*{Acknowledgments}
We are grateful to Martin Gerlach for insightful~suggestions.

\section{Appendices}

\subsection{Metadata, Filtering and Preprocessing} \label{data_description}
The SPGC provides the PG corpus on three different levels of granularity: raw text, time series of word tokens, and~counts of individual words \citep{gerlach2020}. Texts are tokenized using the ``TreebankWordTokenizer'' from the Natural Language Toolkit (NLTK) \citep{nltk2009}. Only tokens consisting of alphabetic characters are kept, meaning that words containing numbers or other symbols are removed. While this processing removes any mentions of numerical objects such as years or ages, it is done to ensure that page and chapter numbers are not erroneously included. Additionally, all tokens are lowercased, as~this ensures that words capitalised after full stops or within dialogues are not considered different words to their standard lowercase~forms.   

PG also provides useful metadata on the texts in the corpus. These metadata provide the title and language of the book, as~well as the author's name, their year of birth, and~their year of death. It also indicates the number of times that a text has been downloaded from the PG website, which in our case was correct as of 18 July 2018. Furthermore, the~metadata contain two sets of manually annotated topical labels for each text: ``subject'' labels and ``bookshelf'' labels. The~subject labels were obtained from the Library of Congress Classification (LCC) or Subject Headings (LCSH) thesauri \citep{egloff2020}, while the bookshelf labels were created by PG volunteers~\cite{gutenberg}. 

For ease of interpretation and application, we decided to only use English texts in our analysis. To~ensure the quality and relevance of the texts used in the analysis, we only included books that were published after 1800, and~that had been downloaded at least \mbox{20 times}. Furthermore, since we wished to use the subject, author's year of birth, and~author's year of death variables in our analysis, any rows with null values in these columns were removed. The~resulting corpus consisted of 13,524 books, containing approximately \mbox{$1.05 \times 10^9$ word~tokens}.

\subsection{Connecting \boldmath{$P(X < Y)$} to ROC Curves} \label{roc_appendix}
We provide a useful link between our performance measure $P(X < Y)$ and receiver operating characteristic (ROC) curves. To~see this, we consider the task of separating the distributions $X$ and $Y$ as described earlier but~reformulate it as a binary classification problem. Given a pair of texts, we compute the dissimilarity and~use a threshold to predict whether those texts come from the same group or different groups. If~the score is below the threshold, we classify it as a within-group value, and~if it is above, we label it~a between-group value. 

If the dissimilarity measure was able to perfectly differentiate between within-group pairwise comparisons and between-group comparisons, then all $X$ values would be smaller than all $Y$ values, and~the distributions of $X$ and $Y$ would not overlap. In~this optimal case, the~threshold for classification would lie between the two distributions. However, when our measure cannot perfectly discriminate between the two distributions, they overlap, and~the optimal threshold value is not obvious. An ROC analysis can be used to not only locate this optimal threshold but~also provide an overall performance evaluation across a range of thresholds. In~an ROC analysis, we iterate over possible threshold values, and~for each one compute the sensitivity (true positive rate, or~TPR) and specificity (true negative rate, or~TNR) of the resulting classification model. In~our case, the~``positives'' were the within-group values $X$, and~the ``negatives'' were the between-group values $Y$. 

ROC curves plot TPR against $1-$TNR and~convey the trade-off between TPR and TNR for different choices of the threshold value. The~shape and position of the curve indicates the overall performance of the metric---the closer the curve is to the top-left corner, the~better the metric is at discriminating between the within-group and between-group dissimilarity values. The~line TPR $=1-$TNR represents the baseline model, whereby we are simply flipping a coin to decide whether to classify samples as a within-group or~between-group value.

The overall performance of the classification model can be quantified by computing the area under the ROC curve (AUC). An~area of 1 represents perfect separation between $X$ and $Y$, while 0.5 is the expected area if the distributions overlapped~completely.  

To establish the connection between $P(X<Y)$ and the AUC, we note the equivalence of the AUC and the Mann--Whitney $U$ test statistic~\cite{mann1947}. The~Mann--Whitney U test is a nonparametric test of the null hypothesis that, for~randomly selected values $X$ and $Y$ from two populations,

\begin{equation*}
    P(X < Y) = P(Y < X).
\end{equation*}
Let $X_1, ..., X_n$ be an i.i.d.~sample from $X$, and~$Y_1, ..., Y_m$ an i.i.d.~sample from $Y$. The~corresponding Mann--Whitney U statistic is defined as
\begin{equation} \label{mann whitney}
    U = \sum_{i=1}^n \sum_{j=1}^m S(X_i, Y_j),
\end{equation}
where
\begin{equation} \label{mann whitney piecewise}
    S(X,Y) = 
    \begin{cases}
    1, & \textrm{if } X < Y, \\
    \frac{1}{2}, & \textrm{if } X = Y, \\
    0, & \textrm{if } X > Y.
    \end{cases}
\end{equation}
As stated earlier, we have that $X$ is a random variable representing the pairwise within-group dissimilarity values, and~$Y$ denotes the pairwise between-group dissimilarity values. Since these values are continuous, we have $P(X = Y) = 0$, so we may ignore the $X = Y$ case in the piecewise function $S(X,Y)$ in Equation~(\ref{mann whitney piecewise}). Thus, the~Mann--Whitney U statistic in Equation~(\ref{mann whitney}) can be interpreted as the number of instances where $X_i < Y_j$ for $i = 1, ..., n$, $j = 1, ..., m$. 

Now, it can be shown that \citep{mason2002, hanley1982}

\begin{equation*}
    AUC = \frac{U}{nm}.
\end{equation*}
Thus, since there are $nm$ total pairings of $X_i$ and $Y_j$ values, the~AUC can be interpreted as the proportion of sample pairings where $X_i < Y_j$ for $i = 1, ..., n$, $j = 1, ..., m$. Hence, the~AUC represents the probability that $X < Y$ given a random within-group dissimilarity score $X$ and a random between-group dissimilarity score $Y$, i.e.,~$P(X < Y)$.

\subsection{Derivation of the Jaccard Distance Lower Bound (Equation~(\ref{jaccard_bound}))} \label{jaccard_appendix}

Let $V_{\bm{\hat{p}}} = |S_{\bm{\hat{p}}}|$ and $V_{\bm{\hat{q}}} = |S_{\bm{\hat{q}}}|$ denote the respective sizes of the vocabularies $S_{\bm{\hat{p}}}$ and $S_{\bm{\hat{p}}}$. Using Heaps' law, we have that
\begin{equation} \label{heaps hn}
    V_{\bm{\hat{p}}} \sim N_{\bm{\hat{p}}}^\beta = h^\beta N^\beta , \quad \quad V_{\bm{\hat{q}}} \sim N_{\bm{\hat{q}}}^\beta = N^\beta.
\end{equation}
Since $h>1$, we conclude that $V_{\bm{\hat{p}}} > V_{\bm{\hat{q}}}$, i.e.,~that $|S_{\bm{\hat{p}}}| > |S_{\bm{\hat{q}}}|$. Thus, since $|S_{\bm{\hat{q}}}| \geq |S_{\bm{\hat{p}}} \cap S_{\bm{\hat{q}}}|$ and  $|S_{\bm{\hat{p}}}| \leq |S_{\bm{\hat{p}}} \cup S_{\bm{\hat{q}}}|$, we~have
\begin{align*}
    |S_{\bm{\hat{q}}}| \geq |S_{\bm{\hat{p}}} \cap S_{\bm{\hat{q}}}| & \implies -|S_{\bm{\hat{q}}}| \leq -|S_{\bm{\hat{p}}} \cap S_{\bm{\hat{q}}}| \\
    & \implies |S_{\bm{\hat{p}}}| - |S_{\bm{\hat{q}}}| \leq |S_{\bm{\hat{p}}} \cup S_{\bm{\hat{q}}}| - |S_{\bm{\hat{p}}} \cap S_{\bm{\hat{q}}}| ,
\end{align*}
Therefore, since $|S_{\bm{\hat{p}}} \cap S_{\bm{\hat{q}}}| \geq 0$, we~have
\begin{equation*}
    |S_{\bm{\hat{p}}}| - |S_{\bm{\hat{q}}}| \leq |S_{\bm{\hat{p}}} \cup S_{\bm{\hat{q}}}| - |S_{\bm{\hat{p}}} \cap S_{\bm{\hat{q}}}| \leq |S_{\bm{\hat{p}}} \cup S_{\bm{\hat{q}}}|.
\end{equation*}
Dividing by $|S_{\bm{\hat{p}}} \cup S_{\bm{\hat{q}}}|$ gives
\begin{equation*}
    \frac{|S_{\bm{\hat{p}}}| - |S_{\bm{\hat{q}}}|}{|S_{\bm{\hat{p}}} \cup S_{\bm{\hat{q}}}|} \leq D_J(S_{\bm{\hat{p}}}, S_{\bm{\hat{q}}}) \leq 1.
\end{equation*}
We can use the relation $|S_{\bm{\hat{p}}} \cup S_{\bm{\hat{q}}}| \leq |S_{\bm{\hat{p}}}| + |S_{\bm{\hat{q}}}|$ to create a more relaxed but also more practical lower bound:

\begin{equation*}
    \frac{|S_{\bm{\hat{p}}}| - |S_{\bm{\hat{q}}}|}{|S_{\bm{\hat{p}}}| + |S_{\bm{\hat{q}}}|} \leq D_J(S_{\bm{\hat{p}}}, S_{\bm{\hat{q}}}) \leq 1.
\end{equation*}
By substituting Equation~(\ref{heaps hn}) and cancelling the common factor $N^\beta$, we~obtain

\begin{equation*}
    \frac{V_{\bm{\hat{p}}} - V_{\bm{\hat{q}}}}{V_{\bm{\hat{p}}} + V_{\bm{\hat{q}}}} \leq D_J(S_{\bm{\hat{p}}}, S_{\bm{\hat{q}}}) \leq 1 \implies \frac{h^\beta - 1}{h^\beta + 1} \leq D_J(S_{\bm{\hat{p}}}, S_{\bm{\hat{q}}}) \leq 1.
\end{equation*}

\subsection{Derivation of the Bias of the Jensen--Shannon Divergence (Table~\ref{jsd_tab})} \label{jsd_appendix}

We say that $\bm{\hat{p}} = (\hat{p}_1, \hat{p}_2,...,\hat{p}_{V_{\bm{\hat{p}}}})$ is the estimated word frequency distribution based on sample $\bm{\hat{p}}$, while $\bm{P} = (p_1, p_2, ...)$ is the true distribution of words in $\bm{P}$. We approximate $H_\alpha(\bm{\hat{p}})$, an~estimator of $H_\alpha(\bm{P})$, using its second-order Taylor expansion around the true probabilities $p_i$:

\begin{align*}
    H_\alpha(\bm{\hat{p}}) \approx & H_\alpha(\bm{p}) + \sum_{i:\hat{p}_i>0}(\hat{p}_i - p_i)\frac{\alpha}{1-\alpha} p_i^{\alpha-1} \\ & - \frac{1}{2}\sum_{i:\hat{p}_i>0}(\hat{p}_i - p_i)^2 \alpha p_i^{\alpha-2},
\end{align*}
where we used that $\frac{\partial H_\alpha}{\partial p_i} = \alpha / (1-\alpha)p_i^{\alpha-1}$ and $\frac{\partial^2 H_\alpha}{\partial p_i \partial p_j} = -\alpha p_i^{\alpha -2}\delta_{i,j}$. 

We can then calculate $\mathbb{E}[H_\alpha(\bm{\hat{p}})]$ by averaging over the different realisations of the random variables $\hat{p}_i$. Here, we assume that the absolute frequency of each word $i$ is drawn from an independent binomial with probability $p_i$ such that $\mathbf{E}[\hat{p}_i] = p_i$ and $\mathbb{V}[\hat{p}_i] = p_i(1-p_i)/N \approx p_i/N$. \cite{martinphd} shows that this yields
\begin{equation} \label{expected entropy}
    \mathbb{E}[H_\alpha(\bm{\hat{p}})] = \frac{1}{1-\alpha} \left( V_{\hat{\bm{p}}}^{(\alpha+1)} - 1 \right) - \frac{\alpha}{2N_{\bm{\hat{p}}}} V_{\hat{\bm{p}}}^{(\alpha)}.
\end{equation}
where $V^{(\alpha)}_{\bm{\hat{p}}}$ denotes the vocabulary size of order $\alpha$,
\begin{equation} \label{vocab order alpha}
    V^{(\alpha)}_{\bm{\hat{p}}} = \sum_{i \in \langle V_{\bm{\hat{p}}} \rangle}p_i^{\alpha - 1}.
\end{equation}
Here, the~notation $\sum_{i \in \langle V_{\bm{\hat{p}}} \rangle}$ indicates that we are summing only over the expected number of observed words $\langle V_{\bm{\hat{p}}} \rangle$ in samples $\bm{\hat{p}}$. 

We are particularly interested in the dependence of $V^{(\alpha)}_{\bm{\hat{p}}}$ on $N_{\bm{\hat{p}}}$. Equation~(\ref{vocab order alpha}) indicates that if we sample $N_{\bm{\hat{p}}}$ words, we expect to observe $V_{\bm{\hat{p}}} = V_{\bm{\hat{p}}}(N_{\bm{\hat{p}}}) \equiv   V^{(\alpha=1)}_{\bm{\hat{p}}}$ unique words. Using the fact that $V_{\bm{\hat{p}}}$ grows as $N_{\bm{\hat{p}}}^{1/\gamma}$ (Heaps' Law), Ref.~\cite{martinphd} shows that $V^{(\alpha)}_{\bm{\hat{p}}}$ scales for large $N_{\bm{\hat{p}}}$ as
\begin{equation}\label{vocab alpha scaling}
    V_{\bm{\hat{p}}}^{(\alpha)} \propto 
    \begin{cases}
    N_{\bm{\hat{p}}}^{-\alpha + 1 + 1/\gamma}, & \alpha < 1 + 1/\gamma \\
    \textrm{constant}, & \alpha > 1 + 1/\gamma .
    \end{cases}
\end{equation}
Here, $\gamma > 1$ is the Zipf exponent (recall that $f(r) \sim r^{-\gamma}$), and~$\alpha$ is the order of the generalised~entropy. 

Since $D_\alpha$ is a linear combination of entropies, we can use Equation~(\ref{expected entropy}) to determine the expected value of its estimator. Suppose we have two samples $\bm{\hat{p}}$ and $\bm{\hat{q}}$ of sizes $N_{\bm{\hat{p}}}$ and $N_{\bm{\hat{q}}}$, respectively. If~we introduce the notation $\bm{\hat{p}\hat{q}} = \frac{1}{2}(\bm{\hat{p}} + \bm{\hat{q}})$, the~expected generalised JSD~is

\vspace{-12pt}
\begin{align*}
    \mathbb{E}[D_{\alpha}(\bm{\hat{p}},\bm{\hat{q}})] = & \; \mathbb{E}[H_\alpha(\bm{\hat{p}\hat{q}})] - \frac{1}{2}\mathbb{E}[H_\alpha(\bm{\hat{p}})] - \frac{1}{2}\mathbb{E}[H_\alpha(\bm{\hat{q}})] \\
    = & \; \frac{1}{1-\alpha} \left\{ V_{\hat{\bm{p}}\hat{\bm{q}}}^{(\alpha + 1)} - \frac{1}{2}V_{\hat{\bm{p}}}^{(\alpha+1)} - \frac{1}{2} V_{\hat{\bm{q}}}^{(\alpha+1)} \right\} \\ & \; + \frac{\alpha}{2N_{\bm{\hat{q}}}}\left( \frac{1}{2} V_{\hat{\bm{p}}}^{(\alpha)} \right) + \frac{\alpha}{2N_{\bm{\hat{q}}}}\left( \frac{1}{2} V_{\hat{\bm{q}}}^{(\alpha)} \right) \\ &  - \frac{\alpha}{2(N_{\bm{\hat{p}}} + N_{\bm{\hat{q}}})} V_{\hat{\bm{p}}\hat{\bm{q}}}^{(\alpha)},
\end{align*}
where $V_{\hat{\bm{p}}\hat{\bm{q}}}^{(\alpha)}$ denotes the generalised vocabulary of order $\alpha$ (Equation~(\ref{vocab order alpha})) for the combined sequence $\bm{\hat{p}\hat{q}} = \frac{1}{2}(\bm{\hat{p}} + \bm{\hat{q}})$, which is of length $N_{\bm{\hat{p}}} + N_{\bm{\hat{q}}}$.

By noting that $V^{(\alpha+1)}_{\bm{\hat{p}}} = \sum_{i \in \langle V_{\bm{\hat{p}}}\rangle}p_i^{\alpha}$, we see that for large $N_{\bm{\hat{p}}}, N_{\bm{\hat{q}}}$,

\begin{align*}
    & \mathbb{E}[D_{\alpha}(\bm{\hat{p}},\bm{\hat{q}})] =   \mathbb{E}[D_{\alpha}(\bm{p},\bm{Q})]  \\ & + \left[\frac{\alpha}{2N_{\bm{\hat{p}}}}\left( \frac{1}{2} V_{\hat{\bm{p}}}^{(\alpha)} \right) + \frac{\alpha}{2N_{\bm{\hat{q}}}}\left( \frac{1}{2} V_{\hat{\bm{q}}}^{(\alpha)} \right) - \frac{\alpha}{2(N_{\bm{\hat{p}}} + N_{\bm{\hat{q}}})} V_{\hat{\bm{p}}\hat{\bm{q}}}^{(\alpha)}\right],
\end{align*}
indicating that the estimator $D_{\alpha}(\bm{\hat{p}},\bm{\hat{q}})$ is biased. We now wish to study this bias, which we define as
\begin{align} \label{jsd bias}
    \textrm{Bias}[D_\alpha(\bm{\hat{p}},\bm{\hat{q}})] = & \frac{\alpha}{2N_{\bm{\hat{p}}}}\left( \frac{1}{2} V_{\hat{\bm{p}}}^{(\alpha)} \right) + \frac{\alpha}{2N_{\bm{\hat{q}}}}\left( \frac{1}{2} V_{\hat{\bm{q}}}^{(\alpha)} \right) \\ & - \frac{\alpha}{2(N_{\bm{\hat{p}}} + N_{\bm{\hat{q}}})} V_{\hat{\bm{p}}\hat{\bm{q}}}^{(\alpha)}.
\end{align}
As we defined in Section~\ref{sec.analytical}, let $N_{\bm{\hat{q}}} = N$ and $N_{\bm{\hat{p}}} = hN$, where $h > 1$. We examine the behaviour of $\textrm{Bias}[D_\alpha(\bm{\hat{p}},\bm{\hat{q}})]$ for a large $N$ by considering the two ways that $V_{\bm{\hat{p}}}^{(\alpha)}$ can scale according to Equation~(\ref{vocab order alpha}).

\subsubsection*{Case 1: $\alpha > 1 + 1/\gamma$}

We begin with the case $V_{\bm{\hat{p}}}^{(\alpha)} = c$, where $c$ is a constant. In~this case, the~bias~becomes
\begin{align*}
    \textrm{Bias}[D_\alpha(\bm{\hat{p}},\bm{\hat{q}})] & = \frac{\alpha}{2(hN)}\left( \frac{c}{2} \right) + \frac{\alpha}{2N}\left( \frac{c}{2} \right) - \frac{\alpha}{2(hN + N)} (c) \\ 
    & = \frac{c\alpha}{2N}\left( \frac{1}{2h} + \frac{1}{2} - \frac{1}{h+1} \right).
\end{align*}
Thus, we see that the decay of the bias is $1/N$. Importantly, $\textrm{Bias}[D_\alpha(\bm{\hat{p}},\bm{\hat{q}})] \to 0$ as $N \to \infty$, so $D_{\alpha}(\bm{\hat{p}},\bm{\hat{q}})$ is asymptotically unbiased. A~perhaps unexpected result is the dependence on the constant $h$. For~simplicity, define
\begin{equation*}
    g_\alpha(h) = \frac{1}{2h} + \frac{1}{2} - \frac{1}{h+1}.
\end{equation*}
We see that $g_\alpha(h)$ is a decreasing function of $h$ when $h \in (1,1+\sqrt{2})$, and~is increasing when $h > 1+\sqrt{2}$. Thus, since $g_\alpha(1) = 0.5$ and $g_\alpha(h) \to 0.5$ as $h \to \infty$, we can conclude that $g_\alpha(h) \leq 0.5$ for all $h \geq 1$. Interestingly, for~a fixed $N$ there is an optimal relation between text lengths, $h^* = 1+\sqrt{2}$, for~which the bias is minimised. The~corresponding minimal value is $g_\alpha(h^*) = \sqrt{2} - 1 \approx 0.4142$.

\subsubsection*{Case 2: $\alpha < 1 + 1/\gamma$}

In this case, we have that $V_{\bm{\hat{p}}}^{(\alpha)} = cN_{\bm{\hat{p}}}^{-\alpha + 1 + 1/\gamma}$, where $c$ is again a constant. The~bias now~becomes
\begin{align*}
    \textrm{Bias}[D_\alpha(\bm{\hat{p}},\bm{\hat{q}})] = & \; \frac{\alpha}{2(hN)}\left( \frac{1}{2} (hN)^{-\alpha + 1 + 1/\gamma} \right) + \frac{\alpha}{2N}\left( \frac{1}{2} N^{-\alpha + 1 + 1/\gamma} \right) \\
    & - \frac{\alpha}{2(hN + N)} (N + hN)^{-\alpha + 1 + 1/\gamma} \\
    = & \; \frac{c\alpha}{2}\left( N^{-\alpha + 1/\gamma} \right) \left(\frac{1}{2} h^{-\alpha + 1/\gamma} + \frac{1}{2} - (h+1)^{-\alpha + 1/\gamma}\right).
\end{align*}
When $\alpha < 1/\gamma$, the~bias diverges with $N \to \infty$ or $h \to \infty$, and~thus, the~estimator $D_\alpha(\bm{\hat{p}},\bm{\hat{q}})$ also diverges. When $1/\gamma < \alpha < 1 + 1/\gamma$, a~sublinear decay with $N$ is observed. Hence, $\textrm{Bias}[D_\alpha(\bm{\hat{p}},\bm{\hat{q}})] \to 0$ as $N \to \infty$, so $D_{\alpha}(\bm{\hat{p}},\bm{\hat{q}})$ is again asymptotically unbiased. Similar to before, we define the~function 
\begin{equation*}
    g_\alpha(h) = \frac{1}{2}h^{-\alpha + 1/\gamma} + \frac{1}{2} - (h+1)^{-\alpha + 1/\gamma}.
\end{equation*}
As in Case 1, we see that $g_\alpha(h)$ is decreasing when $h \in (1,h^*)$ and increasing when $h > h^*$, but~in this case,
\begin{equation*}
    h^* = \frac{2^{\frac{\gamma}{1 - \gamma - \alpha \gamma}}}{1 - 2^{\frac{\gamma}{1 - \gamma - \alpha \gamma}}}
\end{equation*}
We also have that $g_\alpha(h) \to 0.5$ as $h \to \infty$.



\end{document}